\documentclass{article}
\usepackage{spconf,amsmath,graphicx, float, titlesec, lipsum, mathtools, placeins, cuted}

\titlespacing*{\subsection}{0pt}{0.5\baselineskip}{0.5\baselineskip}
\titlespacing*{\section}{0pt}{0.7\baselineskip}{0.7\baselineskip}
\title{Feature Augmentation improves Anomalous Change Detection for Human Activity Identification in Synthetic Aperture Radar Imagery}
\name{Hannah J. Murphy$^{\star}$  \qquad Christopher X. Ren $^{\dagger}$ \qquad Matthew T. Calef$^{\star \dagger}$ }

\address{$^{\star}$ New Mexico Institute of Mining and Technology, Socorro, NM, USA\\
    $^{\dagger}$Intelligence and Space Research Division, Los Alamos National Laboratory, Los Alamos, NM, USA \\
    $^{\star \dagger}$ Descartes Labs, Inc. Santa Fe, NM, USA}

\newcommand\blfootnote[1]{%
  \begingroup
  \renewcommand\thefootnote{}\footnote{#1}%
  \addtocounter{footnote}{-1}%
  \endgroup
}
\begin{document}

%
\maketitle

\begin{abstract}
\blfootnote{contact: cren@lanl.gov, LA-UR-18-20166}
Anomalous change detection (ACD) methods separate common, uninteresting changes from rare, significant changes in co-registered images collected at different points in time. In this paper we evaluate methods to improve the performance of ACD in detecting human activity in SAR imagery using outdoor music festivals as a target. Our results show that the low dimensionality of SAR data leads to poor performance of ACD when compared to simpler methods such as image differencing, but augmenting the dimensionality of our input feature space by incorporating local spatial information leads to enhanced performance.
\end{abstract}
\section{Introduction}
\label{sec:intro}
Quantifying and detecting human activity in remote sensing imagery has important applications in the coordination of humanitarian relief in cases of natural disasters or conflict situations and can provide crucial insights into conditions on the ground. In recent years, many machine learning applications have been developed for the automation of this problem however these applications often require large amounts of labelled data \cite{quinn2018humanitarian}.\par In this work we investigate the applicability of an unsupervised method, anomalous change detection (ACD), to this issue. We also investigate how feature augmentation methods can improve the performance of ACD on this task. Although ACD methods have been applied to optical \cite{TheilerPerkins:2006} and multi-modal imagery \cite{ziemann_multi-sensor_2019}, this work seeks to evaluate their effectiveness for the specific task of identifying human activity within synethatic aperture radar (SAR) imagery. The motivation behind this is that SAR imagery is largely unaffected by weather and cloud occlusion, eliminating many of the pervasive changes which are known to cause variation in optical data, and making it an appealing platform for change detection. Furthermore Sentinel-1, the satellite constellation considered in this work, has a revisit rate of 12 days, providing a large amount of potential data on which to apply change detection. As a proxy for refugee camps, we consider music festivals, in particular Burning Man~\cite{BurningMan:2017} from 2016, Bonnaroo~\cite{Bonnaroo:2017} from 2017, and Coachella~\cite{Coachella:2017} from 2016. We suggest that these outdoor music festivals will have simliar signatures to refugee camps, and as such serve as useful examples of outdoor human activity consisting of temporary buildings and large numbers of occupants. We assess the effectiveness of a number of methods in classifying pixels as changed due to the introduction of the music festival. Our finding is that a particular ACD method, hyperbolic ACD (HACD)~\cite{TheilerPerkins:2006}, when used with an appropriate feature vector, performs as well as, or better than, the other methods we considered.\\

\section{Data}
\label{sect:eval-perf}

\begin{figure*}[!htb]
\begin{minipage}[!htb]{1.0\linewidth}
  \centering
  \centerline{\includegraphics[width=20cm]{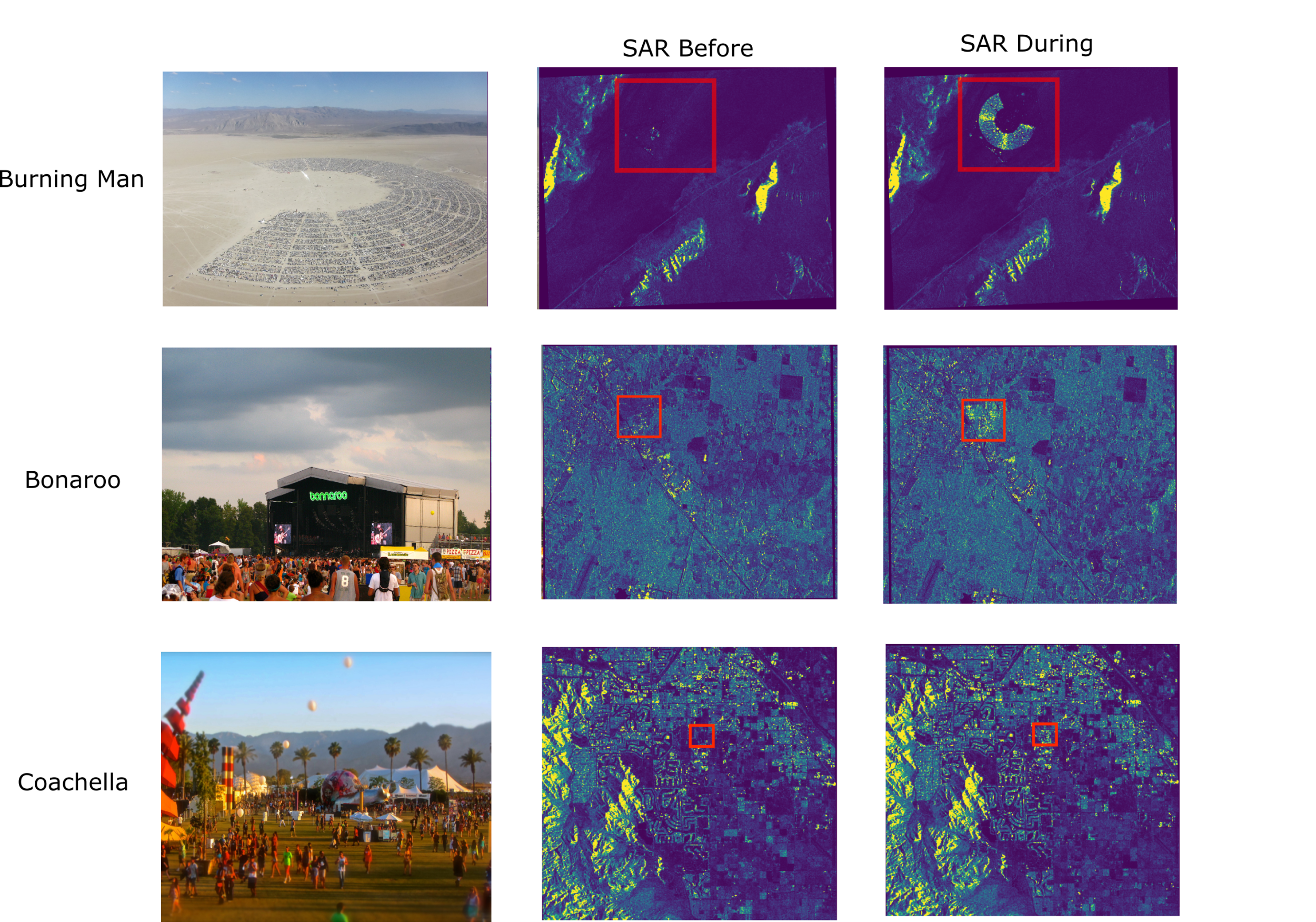}}
  \caption{\label{fig:burning_man}The music festivals studied in this work: Burning Man \cite{BurningMan:2017}, Bonaroo \cite{Bonnaroo:2017} and Coachella \cite{Coachella:2017}, with Sentinel-1 VV images taken before and during the festivals. Red boxes outline areas where human activity of interest takes place in the images.}\medskip
\end{minipage}
\end{figure*}

The Sentinel-1 A and B satellites, launched in the spring of 2014 and the spring of 2016, collect imagery with a roughly fifteen meter ground sample distance (GSD). We use the ground range detected (GRD) imagery from Sentinel-1, and in particular the VV channel. This channel is the vertically polarized component of the reflected signal, where the transmitted signal was also vertically polarized.
The 2016 Burning Man festival took place in the Black Rock Desert of Nevada. The relevant features of the event are that the encampment was large at nearly two and a half miles across, as shown in the left image of the first row of Figure~\ref{fig:burning_man}. It is significant that there was very little change other than the introduction of the encampment. Finally, the background is fairly simple consisting of a valley floor with very little backscatter, and east facing mountains with relatively high backscatter.\par
The 2017 Bonnaroo music festival took place at the Great Stage Park near Manchester Tennessee and is shown in the second row of  Figure~\ref{fig:burning_man} . The festival introduces vehicles, camping, and several large stages, with vehicles were parked in a less structured way than at Burning Man. Unlike Burning Man, the image pair contains some significant coincident changes, in particular changes in backscatter in what appear to be plots of farmland.\par 

The 2016 Coachella music festival takes place at the Empire Polo grounds in Indio, California, as shown in the bottom row of Figure~\ref{fig:burning_man}. While the background changes little between the two images, it is quite varied. The scene includes the Coachella music festivals, a remarkable number of golf courses, residential areas, farmland and mountains. We note that for all three datasets the activity of interest ( true positives) take place within the red bounding boxes, however we apply our change detection algorithms to the entire scenes shown in Figure \ref{fig:burning_man}. 

\section{Experiments}
\label{sect:acd}
\subsection{Performance evaluation}
We use receiver operating characteristic (ROC) curves to assess algorithm performance. In this work the sensitivity parameter is the threshold for the degree of anomalousness assigned to a change at a pixel for that pixel to be marked as part of an anomalous change.
We created masks for the region corresponding to where the music festivals occurred, and used these as our ground truth. Ambiguity in the precise boundaries of the music festivals leads to two masks: the first indicating a region that is entirely within the music festival and a second mask that contains the entire music festival -- in the absence of ambiguity the masks are the same. In such ambiguous cases, we present two ROC curves, and the enclosed band suggest the neighborhood of the true ROC curve.
Our ROC curves are presented as log-log plots, providing a means for the reader to assess low false-positive rate behavior which is the region of interest for ACD algorithms. 

\subsection{Image Differencing}

An intuitive approach to change detection is to subtract two co-registered images pixel-by-pixel. However imagery collected from optical systems can include pervasive differences due to changes in weather or illumination between the two collections. In this case, image differencing can give significant non-zero values that do not reflect changes of interest 
Because the common sources of pervasive changes are largely absent in SAR imagery we use image differencing as a baseline for performance. Throughout the rest of the paper, the image differencing ACD performance baseline will be depicted as blue curve.

\subsection{Hyperbolic ACD}
\label{sub-sect:Overview_HACD}

A central idea behind HACD is to approximate the joint distribution of pixel values in the two images as Gaussian. We shall use $P_{12}$ to denote this distribution, and then $P_{12}(X,Y)$ is the probability density associated with a pixel having value $X$ in the first image, and having value $Y$ in the second. We shall use $P_1$ and $P_2$ to denote the marginal distributions of pixel values for each image. Given these distributions the anomalousness of a pixel change from value $X$ to value $Y$ is assigned the value
\begin{equation}\label{eq:anom}
-\log \frac {P_{12}(X,Y)}{P_1(X)P_2(Y)}.
\end{equation}   
This quantity is high when $P_{12}(X,Y)$ is small and the product of $P_1(X)$ and $P_2(Y)$ is large. In simple terms, anomalousness is high for uncommon changes between common values. By approximating $P_{12}$ as Gaussian, anomalousness becomes quadratic and the coefficients are the covariances. \par
Applying HACD to the per-pixel intensity values in our GRD imagery produces the ROC curves shown in Figure~\ref{fig:acd_patch}a. Interestingly, there is not much benefit to HACD over image differencing. We suggest this is due to a dimensionality issue in the case of SAR data: many optical platforms provide multiple bands per pixel, thereby increasing the dimension of the input space, and providing more opportunities to separate anomalous from pervasive changes.

\subsection{Patch-HACD}
\label{sub-sect:HACD_Patch}

With this as motivation we consider ways to increase the dimension of the input space. As a simple first step we consider a neighborhood around each pixel as input. We use an eleven-by-eleven patch and form a feature vector for each pixel based on the intensity values in the patch centered at that pixel, effectively generating new 'channels' to apply HACD to. We term this approach Patch-HACD. In Figure~\ref{fig:acd_patch}b one can see that, with the exception of the low false-positive regime for Burning Man, this method performs better than image differencing and HACD, indicating that augmenting the feature space can indeed improve the performance of HACD for this application.

\subsection{Gray-level Co-occurrence Matrix-HACD}
\label{sub-sect:HACD_GLCM}

It is a reasonable hypothesis that forming a feature vector from a patch, as we did in section~\ref{sub-sect:HACD_Patch}, encodes more information than is necessary to discriminate pervasive changes from changes due to the introduction of the music festival. We thus consider an alternative and more parsimonious encoding for a patch: our hypothesis is that intensity variations in regions containing music festivals are larger than in other regions and as such consider image texture as a method to capture this difference. We consider the gray-level co-occurrence matrix (GLCM) ~\cite{Haralick:1979}, a statistical parametrization of texture which considers how often pairs of pixels with a specified spatial relationship occur with specific values. The GLCM has been used effectively to analyse texture in SAR imagery~\cite{Soh:1999}.

\begin{figure*}[!t]
\begin{minipage}[!htb]{1.0\linewidth}
  \centering
  \centerline{\includegraphics[width=18cm]{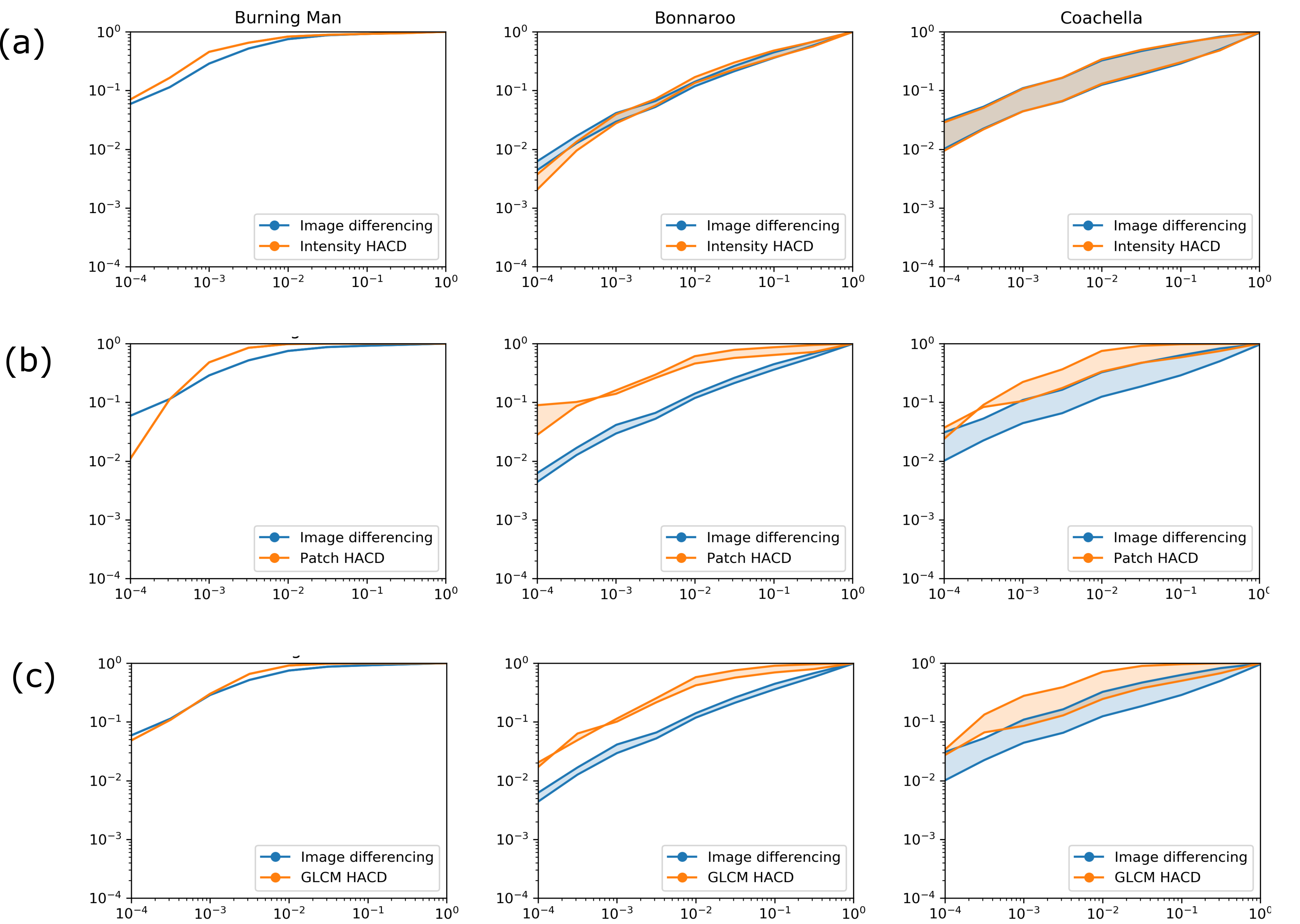}}
\caption 
{ \label{fig:acd_patch}
Log-log ROC curves for: (a) HACD (b) Patch HACD on an 11 by 11 pixel patch around each pixel (c) GLCM - HACD for an 11 by 11 pixel patch around each pixel (d) GLCM - RF for an 11 by 11 pixel patch around each pixel
for the three test cases} 
 \medskip
\end{minipage}
\end{figure*}

Here we form the GLCM for each patch and use that as the per-pixel feature vector. The corresponding ROC curves are shown in Figure~\ref{fig:acd_patch}c.
We see that GLCM-HACD is comparable in performance to Patch-HACD. Performance at Burning Man is markedly better, but performance decreases slightly for the Bonnaroo test case. This may indicate that the texture of the festival is closer to those of the pervasive changes in the scene for the Bonnaroo test case, however it is important to note that there are several parameters that one must choose when forming a GLCM such as the number of discrete intensity levels. As such the results we present here can be considered a lower bound for best performance of GLCM on the test cases we consider, and can potentially be ameliorated through careful hyperparameter search.

\section{Conclusions}
\label{sect:conclusions}

We report that utilising HACD to detect outdoor human activities in our test cases outperformed image differencing only in the cases where dimensionality augmentation such as utilising a local image patch for each pixel or calculating the GLCM for said image patch was used. This suggests that the combination of dimensionality augmentation and ACD algorithms can be a powerful tool for detecting human activity in SAR imagery, a data modality where it is currently not often used. An interesting extension of this study would be to evaluate the performance of these methods for detecting similar changes in multi-modal imagery, i.e combining optical and SAR imagery.  
\bibliographystyle{IEEEbib}
\bibliography{report}

\begin{thebibliography}{1}

\bibitem{quinn2018humanitarian}
John~A Quinn, Marguerite~M Nyhan, Celia Navarro, Davide Coluccia, Lars Bromley,
  and Miguel Luengo-Oroz,
\newblock ``Humanitarian applications of machine learning with remote-sensing
  data: review and case study in refugee settlement mapping,''
\newblock {\em Philosophical Transactions of the Royal Society A: Mathematical,
  Physical and Engineering Sciences}, vol. 376, no. 2128, pp. 20170363, 2018.

\bibitem{TheilerPerkins:2006}
James Theiler and Simon Perkins,
\newblock ``Proposed framework for anomalous change detection,''
\newblock in {\em ICML Workshop on Machine Learning Algorithms for Surveillance
  and Event Detection}, 2006, pp. 7--14.

\bibitem{ziemann_multi-sensor_2019}
Amanda~K. Ziemann, Christopher~X. Ren, and James Theiler,
\newblock ``Multi-sensor anomalous change detection at scale,''
\newblock in {\em Algorithms, {Technologies}, and {Applications} for
  {Multispectral} and {Hyperspectral} {Imagery} {XXV}}, Baltimore, United
  States, May 2019, p.~37, SPIE.

\bibitem{BurningMan:2017}
Wikipedia.org,
\newblock ``{Burning Man},'' 2017.

\bibitem{Bonnaroo:2017}
Wikipedia.org,
\newblock ``{Bonnaroo Music Festival},'' 2017.

\bibitem{Coachella:2017}
Wikipedia.org,
\newblock ``{Coachella Valley Music and Arts Festival},'' 2017.

\bibitem{Haralick:1979}
Robert~M Haralick,
\newblock ``Statistical and structural approaches to texture,''
\newblock {\em Proceedings of the IEEE}, vol. 67, no. 5, pp. 786--804, 1979.

\bibitem{Soh:1999}
L-K Soh and Costas Tsatsoulis,
\newblock ``{Texture analysis of SAR sea ice imagery using gray level
  co-occurrence matrices},''
\newblock {\em IEEE Transactions on geoscience and remote sensing}, vol. 37,
  no. 2, pp. 780--795, 1999.

\end{thebibliography}

\end{document}